\documentclass[fleqn,10pt]{wlscirep}
\usepackage[utf8]{inputenc}
\usepackage[T1]{fontenc}
\usepackage{url}

\title{End-to-end plaque counting and virus titration from laboratory plate images with deep learning}

\author[1,+,*]{Eugenia~Moris}
\author[2,3,4,5,+]{Alicia~Costábile}
\author[2,3]{Sebastián~Rey}
\author[2,3]{Irene Ferreiro}
\author[2,3]{Joaquín~Hurtado}
\author[1]{Lizandra~Lissette~Luciano}
\author[1]{Matías~Villagrán}
\author[1]{Aisha~Espino~Vázquez}
\author[1]{Jomari~Ramos}
\author[1]{Isadora~Monteiro}
\author[1]{María~Victoria~de~Santiago}
\author[2,3,4]{Pilar~Moreno}
\author[2,3,4]{Gonzalo~Moratorio}
\author[1,6,*]{José~Ignacio~Orlando}

\affil[1]{Arionkoder LLC, Boston, MA 02114, United States}

\affil[2]{Laboratory of Experimental Virus Evolution, Pasteur Institute of Montevideo, Montevideo, 11400, Uruguay}
\affil[3]{Laboratory of Molecular Virology, Faculty of Sciences, University of the Republic, Montevideo, 11400, Uruguay}
\affil[4]{Center for Innovation in Epidemiological Surveillance, Pasteur Institute of Montevideo, Montevideo, 11400, Uruguay}
\affil[5]{Biochemistry Section, Faculty of Sciences, University of the Republic, Montevideo, 11400, Uruguay}

\affil[6]{Yatiris Group, PLADEMA Institute, CONICET / UNICEN, Tandil, 7000, Argentina}

\affil[*]{Corresponding author, eugenia.moris@arionkoder.com, ignacioo@arionkoder.com}
\affil[+]{these authors contributed equally to this work}

\begin{abstract}
Plaque assays remain the gold standard readout of virus infectivity; however, plaque counting from plate images is labor-intensive and prone to inter-operator variability. We present an end-to-end, computer-aided workflow for cytopathic effect–based virus titration directly from laboratory plaque assay images. The proposed approach combines two models derived from the Segment Anything Model (SAM): a SAM2-based well-segmentation module that localizes assay wells across heterogeneous imaging conditions, and a SAM-based plaque-segmentation model that detects and enumerates plaques within each well.
The method was evaluated on a mixed dataset comprising private plaque assay images of Mayaro virus and Coxsackievirus B3, together with public Vaccinia virus images from the VACVPlaque dataset. The pipeline outputs per-well plaque counts, automatically computes plaque-forming units per milliliter (PFU/mL), and is integrated into a web-based platform that allows users to review results and organize experiments.
On held-out plates (17 from MAYV/CVB3 and 22 from VACV), the workflow generalized across two plate formats (6-well and 12-well) and showed strong agreement with manual annotations (Pearson correlation coefficients of 0.92 for MAYV/CVB3 and 0.88 for VACV). Automated plaque counts were further compared with annotations from four independent experts, demonstrating high concordance. The proposed system will be open-sourced and publicly released upon acceptance of this manuscript to enable reproducible, scalable, and audit-ready plaque assay analysis while substantially reducing manual annotation effort.

\end{abstract}
\begin{document}

\flushbottom
\maketitle
% * <john.hammersley@gmail.com> 2015-02-09T12:07:31.197Z:
%
%  Click the title above to edit the author information and abstract
%
\thispagestyle{empty}

\section*{Introduction}
Plaque assays is the gold-standard readout of infectious virus, but quantifying plaques from plate images is still routinely performed by visual inspection, which is slow, subjective, and difficult to standardize across analysts and laboratories \cite{Masci2019VirPlaque}. In response, several software tools have emerged to automate parts of the workflow, typically under controlled imaging conditions or with fluorescent readouts. Examples include Plaque2.0, a fluorescence microscopy–based high-content framework \cite{Yakimovich2015Plaque20}, Viridot for immunofocus plaque counting and titer estimation on 96-well plates \cite{Katzelnick2018Viridot}, and Fiji/ImageJ macros and command-line utilities such as ViralPlaque and the Plaque Size Tool \cite{Cacciabue2019ViralPlaque, Trofimova2021PST}. Beyond classical image processing, machine-learning approaches using $k$-means–based segmentation and bespoke hardware have been described for specific viruses and plate formats \cite{Phanomchoeng2022PeerJCS}. Yet many of these solutions require specialized optics (e.g., fluorescence or holography) \cite{Liu2023NatBiomedEng} or laborious parameter tuning, limiting adoption for routine, label-free cytopathic-effect (CPE) plaque assays captured as simple photographs.

Recent advances in foundation-model segmentation offer a path to robust, annotation-efficient analysis under diverse imaging conditions. The Segment Anything Model (SAM) demonstrated strong zero-shot generalization across image domains, enabling promptable segmentation with minimal task-specific labels \cite{Kirillov2023SAM}. In parallel, openly available, annotated datasets of plaque assays imaged via mobile photography--such as VACVPlaque---now make it possible to train and benchmark models on realistic, label-free CPE plates without specialized imaging hardware \cite{De2025VACVPlaque}. Combined, these developments suggest accurate and scalable plaque counting directly from laboratory plate photographs with reduced manual effort and analyst bias.

%Here we introduce an end-to-end method for computer-aided CPE-based virus titration directly from plate images. Our approach pairs two SAM-derived models: a well-segmentation network that localizes wells across plate formats and lighting conditions, and a plaque-segmentation network that detects and enumerates plaques within each well. The pipeline outputs per-well plaque detection and counting, automatically computing plaque-forming units per millilitre (PFU/mL). We integrated these developments into a single web-platform for CPE titration, called Titra\footnote{\url{https://titra.app/}}, featuring the automated machine learning-based sequence but also incorporating an interface for rapid quality control and experiment organization. Relative to prior stand-alone desktop tools \cite{Emi2025PlaQuest} and earlier rule-based methods \cite{Cacciabue2019ViralPlaque, Trofimova2021PST, Katzelnick2018Viridot}, our contributions are threefold: (i) dual SAM-guided models that reduce annotation burden while maintaining accuracy; (ii) an integrated workflow that accelerates routine titration and yields reproducible, audit-ready PFU/mL calculations at scale; and (iii) the first open source and web-native tool for computer-aided CPE-based titration.

We introduce an end-to-end method for computer-aided CPE-based virus titration directly from plate images. Our approach combines two SAM-derived models: a well-segmentation network that localizes wells across plate formats and imaging conditions, and a plaque-segmentation network that detects and counts plaques within each well. The pipeline outputs per-well plaque counts and automatically computes plaque-forming units per millilitre (PFU/mL). We integrate these components into a web platform for CPE titration, Titra (\url{https://titra.app/}), which provides automated ML-based analysis together with rapid quality control and experiment management. Compared with prior stand-alone tools \cite{Emi2025PlaQuest} and rule-based methods \cite{Cacciabue2019ViralPlaque, Trofimova2021PST, Katzelnick2018Viridot}, our contributions are: (i) dual SAM-guided models reducing annotation burden while maintaining accuracy; (ii) an integrated workflow enabling scalable, reproducible PFU/mL estimation; (iii) an open-source, web-native tool for computer-aided CPE-based titration, to be released upon acceptance of this manuscript; and (iv) MAYV/CVB3, a curated dataset of plaque assay images collected at the Institut Pasteur de Montevideo under routine laboratory conditions using smartphone photography (6- and 12-well plates), planned for public release upon acceptance of this manuscript.

\section*{Results}

\subsection*{Overview of the proposed approach}

\begin{figure}[t]
    \centering
    \includegraphics[width=\linewidth]{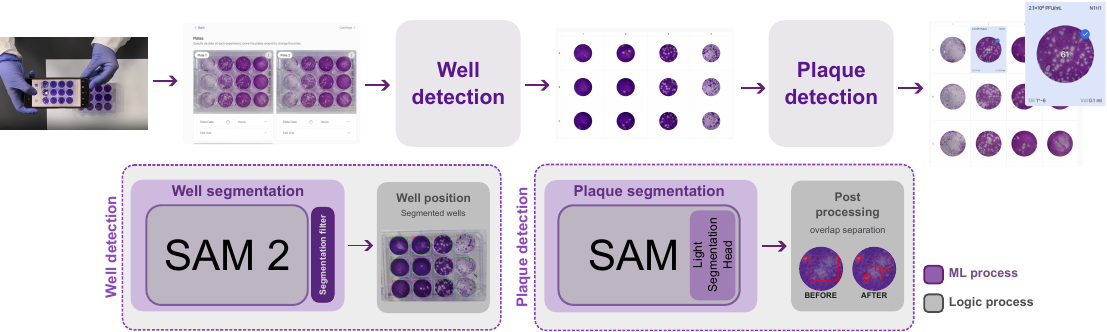}
    \caption{ End-to-end workflow for automated plaque assay quantification in \textit{Titra}. A full-plate image is processed using a SAM2-based well-segmentation module to identify and localize individual wells (Well Detection module). Each segmented well is subsequently analysed by a SAM-based plaque-segmentation model, followed by post-processing to resolve overlapping plaques (Plaque Detection module). The resulting plaque counts are combined with dilution and inoculation information to compute PFU/mL values.}
    \label{fig:schematic}
\end{figure}

Figure~\ref{fig:schematic} illustrates the end-to-end workflow used to compute %plaque-forming units per milliliter 
PFU/mL from a plaque assay plate image. Starting from a full-plate image, the pipeline performs automated well detection, followed by plaque detection within each individual well to enable plaque counting.

The input consists of an image of a plaque assay plate acquired using a standard camera, such as a mobile phone. The image is first processed by the SAM2 mask generator \cite{ravi2024sam} to produce candidate segmentation masks. These masks are subsequently filtered to retain only valid wells. The resulting well segmentation is used both to isolate individual wells and to estimate their spatial arrangement within the plate, allowing each well to be correctly associated with its corresponding dilution position for downstream PFU/mL calculation.

Each segmented well is then processed independently by the plaque-detection stage. The cropped well image is passed to a plaque-segmentation model based on the SAM\cite{kirillov2023segment}, using a frozen image encoder and a lightweight, task-specific decoder trained for plaque segmentation. A dedicated post-processing step is applied to refine plaque boundaries and separate overlapping plaques, yielding a final plaque count for each well.

Finally, the estimated plaque counts, together with the corresponding virus dilution and inoculation volume, are used to compute the PFU/mL value for the assay.

\subsection*{Well detection results}

\begin{figure}[t]
    \centering
    \includegraphics[width=\linewidth]{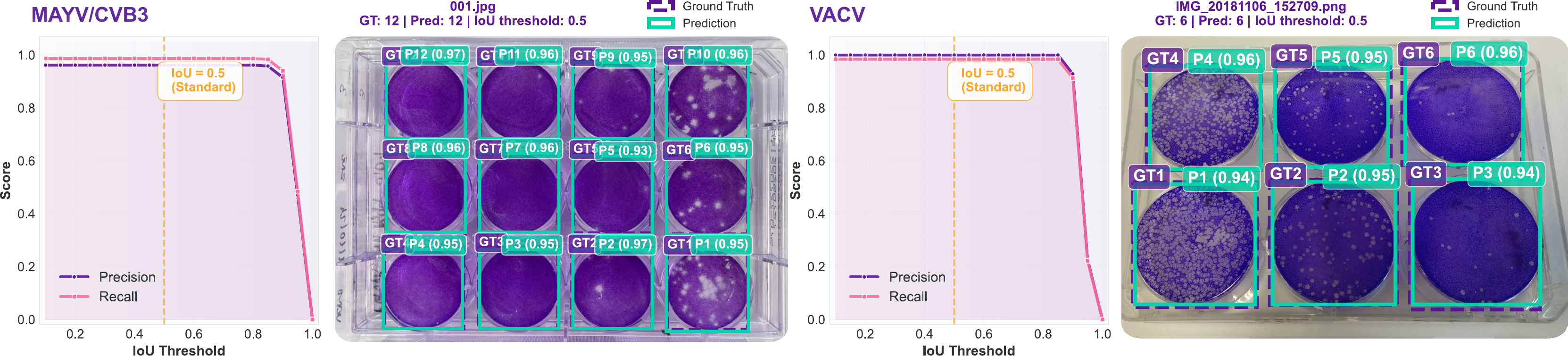}
    \caption{Automated well detection results in MAYV/CVB3 (left) and VACV (right). Left: precision and recall variations for different Intersection over Union (IoU) thresholds. Right: Representative examples of well detections, including ground truth (GT) boxes (purple) and predicted wells (light green).}
    \label{fig:well-segmentation-performance}
\end{figure}

Figure~\ref{fig:well-segmentation-performance} illustrates the performance of the well-segmentation model on the MAYV/CVB3 and VACV datasets. For each dataset, quantitative results are shown on the left, while representative qualitative examples are presented on the right. 

%Numeric evaluation was performed quantifying precision and recall at different Intersection over Union (IoU) thresholds, as typically done when estimating the accuracy of object detection models \cite{nguyen2022trustworthy}. Notice that in both MAYV/CVB3 and VACV datasets, a similar behaviour is observed, with high precision and recall values observed across most of the IoU range, except when considering very high thresholds (above 0.8). For the MAYV/CVB3 dataset, recall is slightly higher than precision across that range (0.9898 and 0.9614, respectively). The opposite behaviour is observed in VACV, were precision is slightly higher than recall (0.9920 vs. 0.9841, respectively). Given this stable behaviour, a standard IoU threshold of 0.5 was fixed for all the remaining experiments.

%Performance was evaluated using precision and recall across different Intersection over Union (IoU) thresholds, following standard object detection practice~\cite{nguyen2022trustworthy}. Both MAYV/CVB3 and VACV exhibit similar behaviour, with high precision and recall across most IoU values, except at strict thresholds ($>0.8$). For MAYV/CVB3, recall is slightly higher than precision (0.9898 vs. 0.9614), whereas VACV shows the opposite trend (0.9920 vs. 0.9841). Given this stability, an IoU threshold of 0.5 was fixed for subsequent experiments.

Performance was evaluated using precision and recall across multiple Intersection over Union (IoU) thresholds, following standard object detection practice~\cite{nguyen2022trustworthy}. MAYV/CVB3 and VACV showed similar behaviour, with high precision and recall across most IoU values except at strict thresholds ($>0.85$). For MAYV/CVB3, recall slightly exceeded precision (0.9898 vs. 0.9614), whereas VACV showed the opposite trend (0.9920 vs. 0.9841). Given this stability, an IoU threshold of 0.5 was used for subsequent experiments.

The qualitative examples in Figure~\ref{fig:well-segmentation-performance} show representative well-segmentation results for each dataset, with ground-truth annotations displayed in violet and model predictions shown in light-green. IoU values for each detection are included for reference. The MAYV/CVB3 example corresponds to a 12-well plate acquired from a near top-down viewpoint under uniform illumination. All wells were detected in this case, with IoU values above 0.93. For VACV, the example shows a 6-well plate acquired from a different viewing angle and under less stable illumination conditions; in this case, all wells were also detected by the model, with IoU values above 0.94.

%Figure~\ref{fig:well-segmentation-not-detected} shows two plate examples where the model failed to identify some wells. On the left is an MAYV/CVB3 plate, where the model missed 3 of the 12 wells, and on the right is a VACV plate, where one of the six wells was not detected. In all those cases we observe that the wells are completely transparent of the plaques consuming almost all the cells.

\begin{figure}[t]
    \centering
    \includegraphics[width=.7\linewidth]{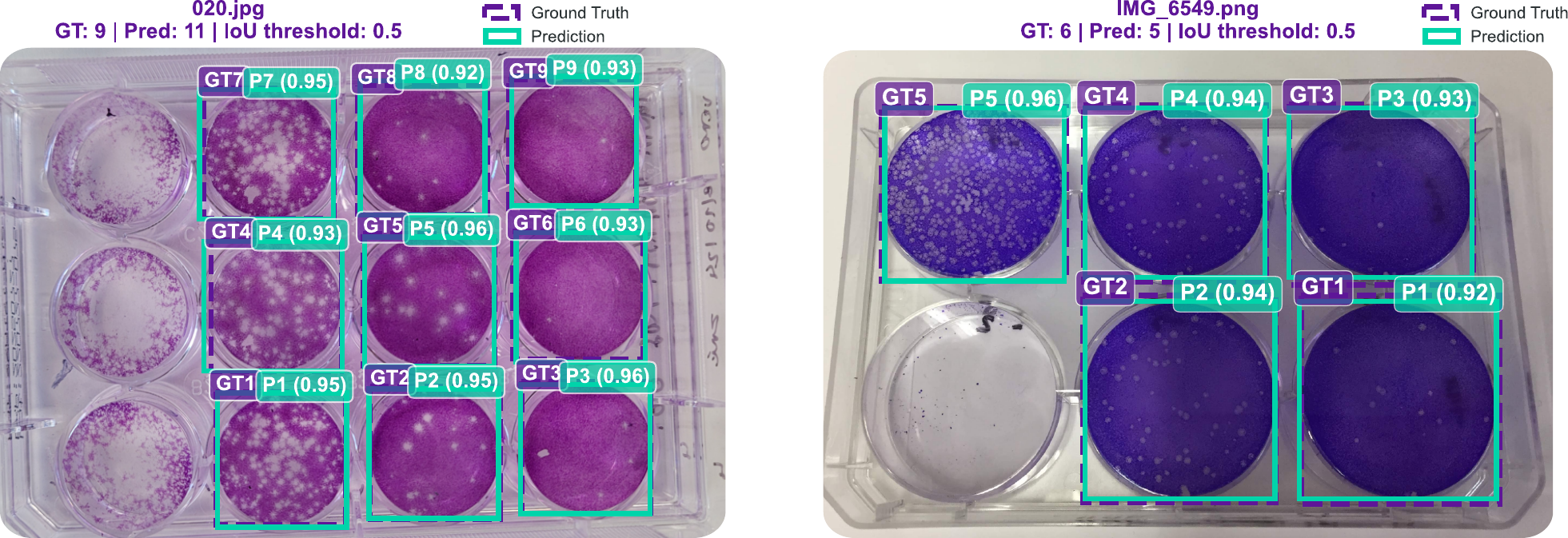}
    \caption{Examples of well-detection failure cases. Left: MAYV/CVB3 plate example where three out of twelve wells were not detected. Right: VACV plate example where one of six wells was missed.}
    \label{fig:well-segmentation-not-detected}
\end{figure}

Figure~\ref{fig:well-segmentation-not-detected} illustrates representative failure cases of the well segmentation stage. In the MAYV/CVB3 example (left), three of the twelve wells were not detected, while in the VACV example (right), one of six wells was missed. In all cases, the undetected wells appear almost completely transparent, exhibiting minimal staining and lacking visible plaque contrast relative to the surrounding plate background.

\subsection*{Plaque detection results}

\begin{figure}[t]
    \centering
    \includegraphics[width=\linewidth]{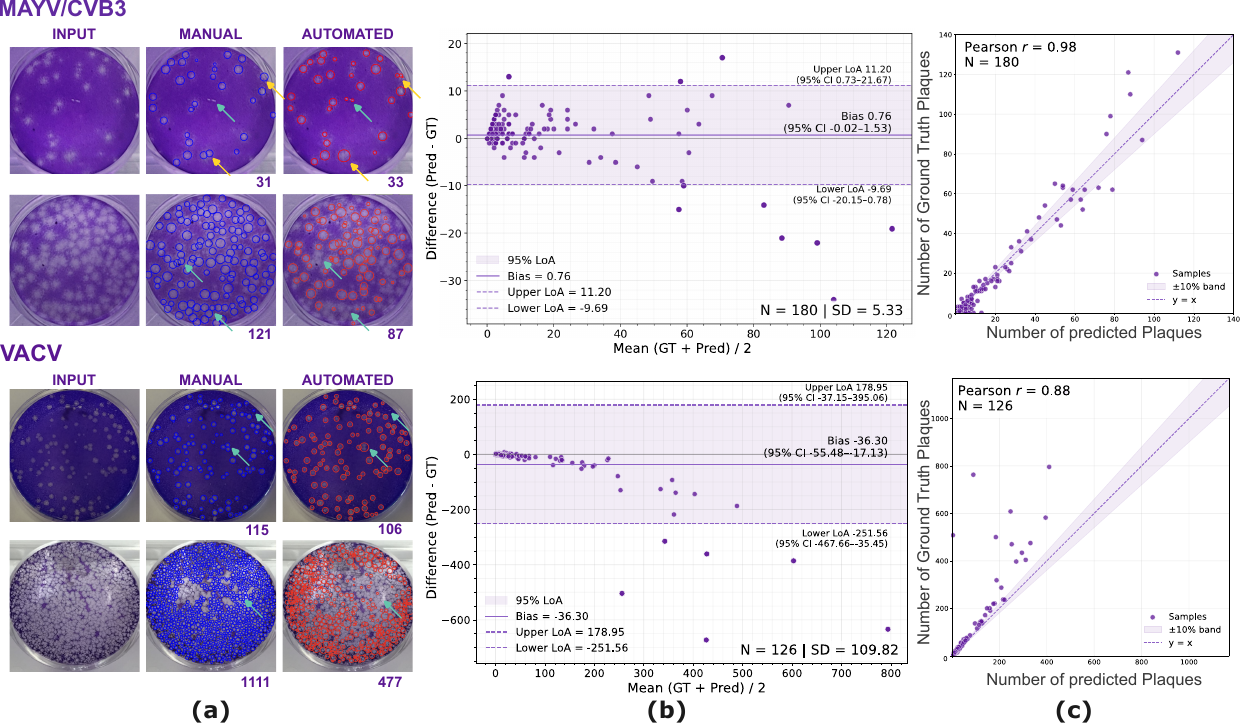}
    \caption{Automated plaque detection results in the MAYV/CVB3 and VACV datasets. (a) Representative plaque assay wells from the MAYV/CVB3 (top) and VACV (bottom) datasets--left to right: input images, manual annotations, and automated segmentation outputs, including plaque counts below; top to bottom: wells with low and high plaque density. (b) Bland–Altman plots showing agreement between automated predictions (Pred) and manual ground truth counts (GT) across all test wells. (c) Correlation plots comparing automated and manual plaque counts, with dashed lines denoting ±10\% deviation bands.}
    \label{fig:plaque-results}
\end{figure}

Figure~\ref{fig:plaque-results} presents the results obtained for plaque detection in the MAYV/CVB3 and VACV datasets.

Figure~\ref{fig:plaque-results}(a) shows qualitative examples of two representative scenarios for each dataset, with different plaque densities. The input image, the manual annotation, and the automated segmentation are shown in each case, together with the corresponding plaque count below.
Overall, the VACV dataset exhibits a higher number of plaques per well compared with MAYV/CVB3, with plaques distributed across the entire well area, including near the borders. In both datasets, overlap between plaques is observed, particularly in images with higher plaque density. Differences in plaque size are also visible between datasets, with VACV plaques appearing smaller than those in MAYV/CVB3.

For MAYV/CVB3, the first row shows a well with a low plaque concentration (31 plaques according to manual segmentation). The automated method detected 33 plaques, with discrepancies occurring in regions with overlapping plaques, such as the bottom-centre area and the right side of the well (yellow arrow), as well as the detection of a small artefact in the centre (light-green arrow). In the second MAYV/CVB3 example, corresponding to a much higher plaque concentration (121 manually annotated plaques), the automated method detected fewer plaques (87). Missed detections are again primarily observed in regions with strong plaque overlap, particularly near the bottom of the well and along the borders, whereas overlapping plaques in the central region are more frequently detected by the method.

For VACV, the first example shows a well with a relatively low plaque concentration (115 plaques manually segmented), although this number is substantially higher than that observed in typical MAYV/CVB3 examples. The automated method detected 106 plaques, with differences mainly associated with overlapping plaques in the central region of the well and a smaller plaque on the upper side of the well. In the second VACV example, which has a very high plaque concentration (1111 plaques manually annotated), plaques cover most of the well area. The manual annotation shows reduced plaque delineation in the central region of this well. In this case, the automated method detected only 477 plaques, missing several plaques located in the central region. Conversely, the model accurately detected most of the plaques near the well borders.

%Figure~\ref{fig:plaque-results}(b) presents the Bland–Altman analysis comparing predicted and ground truth plaque counts for the MAYV/CVB3 (upper) and VACV (lower) datasets. The central horizontal line indicates the mean difference (bias), which was $0.76$ for MAYV/CVB3 and $-36.30$ for VACV. The upper and lower limits of agreement were $11.20$ and $-9.69$ for MAYV/CVB3, and $178.95$ and $-251.56$ for VACV, respectively. For both datasets, most data points fall within the limits of agreement. In the VACV dataset, the differences tend to become more negative as the mean plaque count increases, indicating under-detection when plaque density is substantially higher (as in the examples shown in Figure~\ref{fig:plaque-results}(a)). In contrast, the MAYV/CVB3 dataset shows a more scattered pattern as the number of plaques increases.

Figure~\ref{fig:plaque-results}(b) shows the Bland–Altman analysis comparing predicted and ground truth (GT) plaque counts for the MAYV/CVB3 (upper) and VACV (lower) datasets. The mean difference (bias) was $0.76$ for MAYV/CVB3 and $-36.30$ for VACV. The corresponding limits of agreement were $11.20$ and $-9.69$ for MAYV/CVB3, and $178.95$ and $-251.56$ for VACV. In both datasets, most points fall within these limits. For VACV, differences become increasingly negative at higher mean plaque counts, indicating under-detection in high-density wells (Figure~\ref{fig:plaque-results}(a)). In contrast, MAYV/CVB3 displays a more dispersed pattern as plaque counts increase.

Figure~\ref{fig:plaque-results}(c) shows scatter plots comparing the number of predicted plaques with the GT counts for the MAYV/CVB3 and VACV datasets. For MAYV/CVB3, a strong linear correlation is observed between predicted and GT counts, with a Pearson correlation coefficient of $0.98$ ($N = 180$). For the VACV dataset, the overall Pearson correlation coefficient is lower, at $0.88$ ($N = 126$), largely driven by cases with a larger number of plaques.

\begin{figure}
    \centering
    \includegraphics[width=\linewidth]{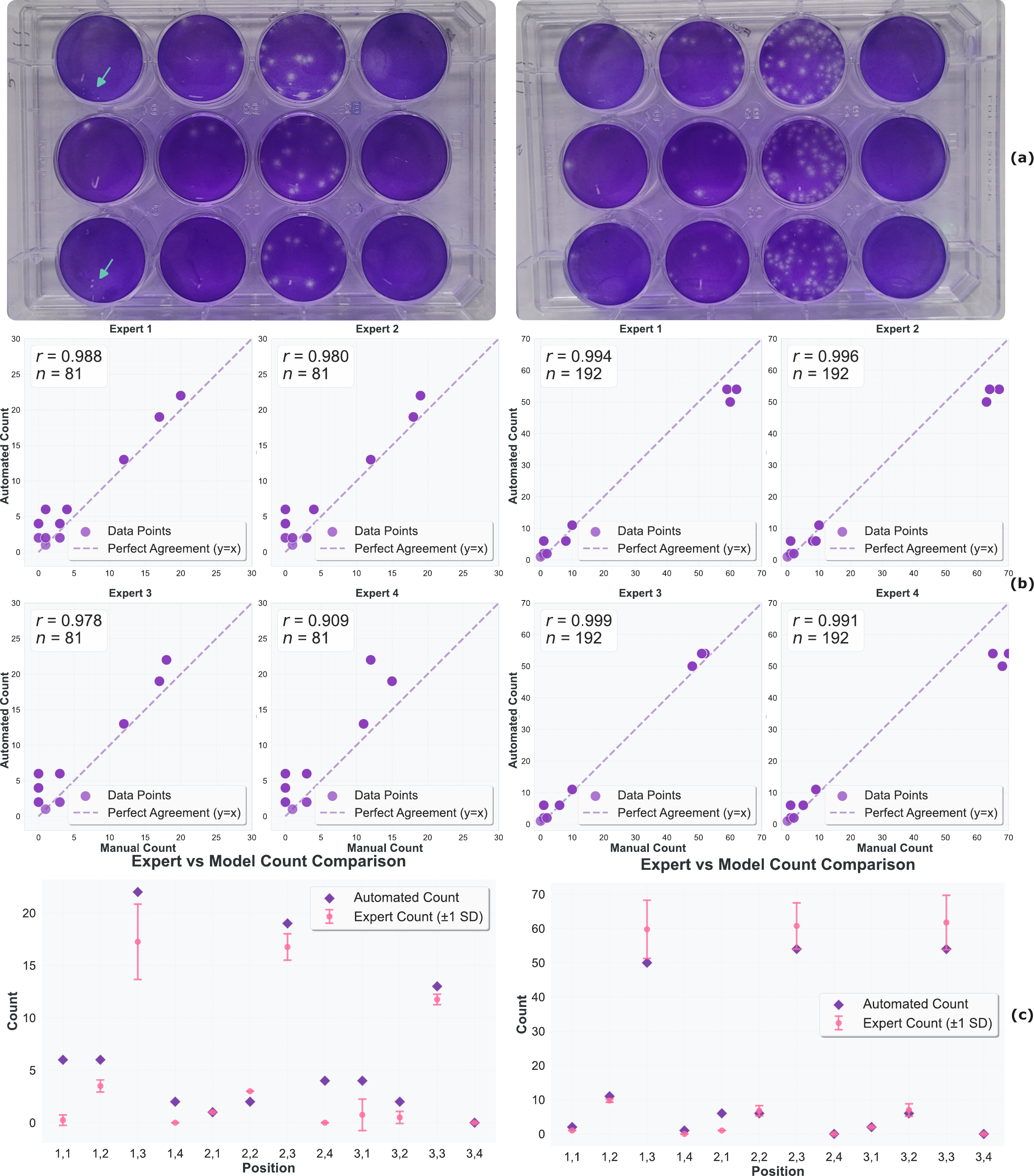}
    \caption{Results for two MAYV/CVB3 plaque count samples. Comparison between automated plaque counts and manual plaque counts performed by four experts. (a) Representative plaque assay plates used for evaluation. (b) Plaque counts from two samples (1, left; 2, right) obtained using the automated segmentation method and by four independent human experts (Experts 1–4). Scatter plots show the correlation between automated and individual expert counts for each plate, together with the corresponding Pearson correlation coefficients. (c) Comparison between automated plaque counts (violet diamond) and the mean $\pm$ 1 SD of the four expert counts (pink) for each well position.}
    \label{fig:comparison-with-experts}
\end{figure}

To study these results in the context of interobserver variability, Figure~\ref{fig:comparison-with-experts} compares plaque counts obtained using the proposed workflow with those provided by four independent experts for the two plaque assay plates from the MAYV/CVB3 dataset shown at the top of the figure, each exhibiting different plaque concentrations. The middle panels present scatter plots comparing automated plaque counts with those reported by each expert (Experts~1–4) for both samples. Each scatter plot shows the comparison between our approach and each expert individually, together with the corresponding Pearson correlation coefficient. The bottom panels display, for each well position (row~$\times$~column), the automated plaque count (violet diamond) together with the mean and standard deviation (mean~$\pm$~SD) of the expert counts.

The plate on the left exhibits low overall plaque counts. Pearson correlation coefficients between automated and expert counts are high, ranging from 0.909 to 0.988. The largest differences are observed in wells with lower plaque numbers, whereas wells with higher plaque counts (between 10--30 plaques) show closer agreement with the experts overall; only Expert~4 appears to differ more noticeably. The per-position comparison shows that automated counts are generally higher than the expert mean, with positions (2,1), (2,2), and (3,4) showing equal or lower automated counts relative to the experts. Position (1,1) shows the largest difference between automated and expert counts. Furthermore, greater inter-expert variability is observed in wells with higher plaque counts. Only position (3,1) shows notable variability among experts at low plaque counts. A detailed inspection of the samples shows that the automated method detects small plaques that are not consistently annotated by the experts, as illustrated at positions (1,1) and (3,1) with light-green arrows.

%A similar behaviour is observed in the case illustrated on the right-hand side of Figure~\ref{fig:comparison-with-experts}. Strong linear correlations are seen (values ranging from 0.996 to 0.998). The largest differences between automated and expert counts are observed in wells with higher plaque numbers (approximately 50--70 plaques) for three of the four experts, with Expert~3 showing the closest agreement. Additionally, there is one position in which the automated count is higher than that reported by the experts. When comparing counts by well position, the largest discrepancy is observed at position (2,1), whereas for the remaining positions the automated counts fall within the variability range of the expert annotations. Increased variability among expert counts is also observed in wells with higher plaque numbers, while almost no discrepancies are seen when lower numbers of plaques are present.

A similar behaviour is observed in the right-hand panel of Figure~\ref{fig:comparison-with-experts}, where strong linear correlations are obtained ($r=0.994$–$0.999$). The largest discrepancies between automated and expert counts occur in wells with higher plaque numbers (approximately 50–70 plaques), affecting three of the four experts, with Expert~3 showing the closest agreement. One well position exhibits a higher automated count than the expert consensus. When analysed by position, the greatest difference appears at (2,1), whereas counts at the remaining positions fall within the inter-expert variability range. Notably, variability among experts increases with plaque density, while wells with lower plaque numbers show minimal disagreement.

\begin{figure}
    \centering
    \includegraphics[width=0.7\linewidth]{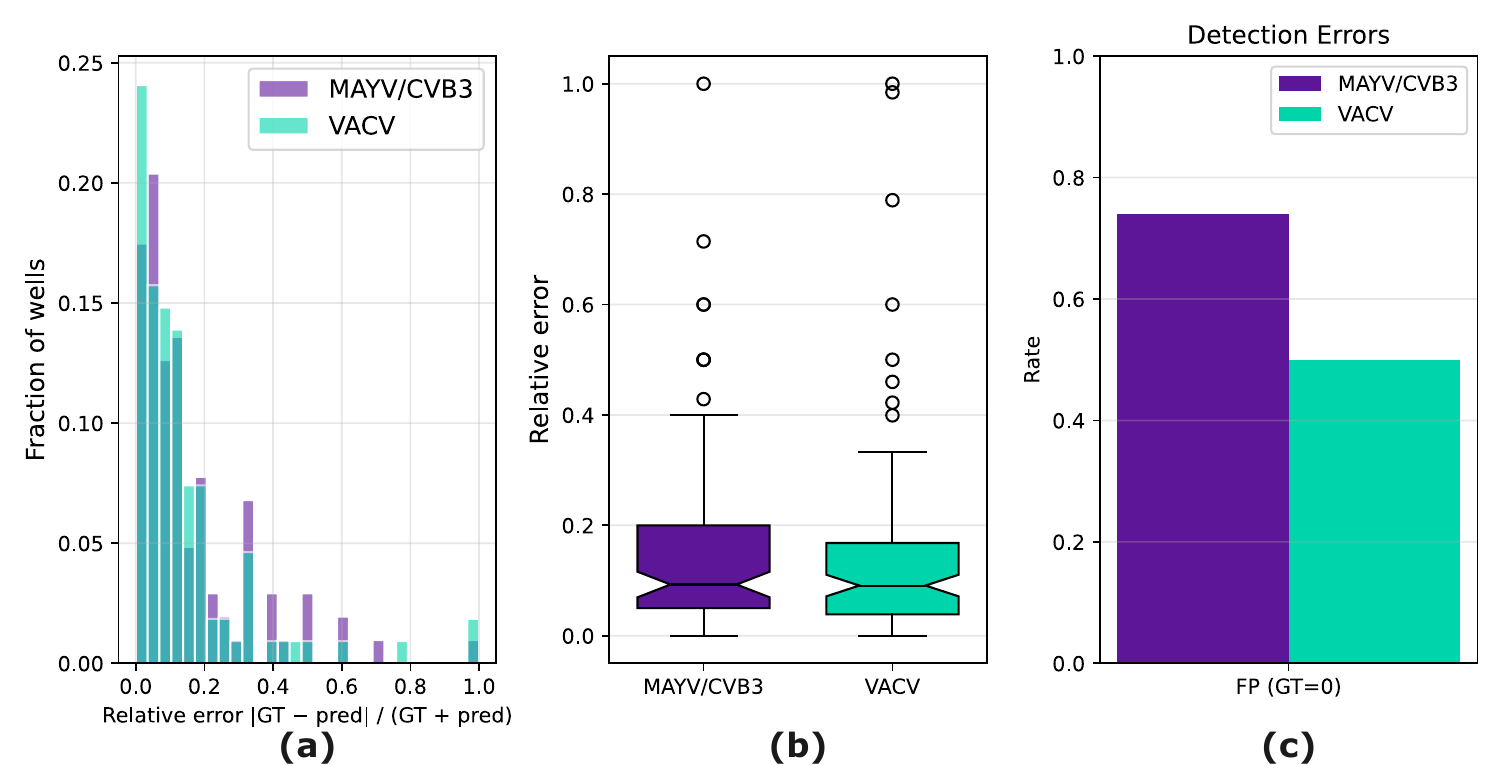}
    \caption{Symmetric relative plaque counting error and false positives in empty wells for MAYV/CVB3 and VACV. a: distribution of |GT - pred|/(GT + pred) for wells with plaques (GT > 0). b: boxplot summary. c: false-positive rate in empty wells (GT = 0). MAYV/CVB3 (purple), VACV (green).}
    \label{fig:error}
\end{figure}

%To further characterise the trends and differences in plaque-count errors between datasets, Figure~\ref{fig:error} presents a histogram (left) and a boxplot (right) illustrating the distribution of differences between ground truth and predicted plaque counts, normalised by the ground truth count for each well. Positive values indicate a higher number of false negatives, whereas negative values indicate a higher number of false positives. A partial overlap is observed between datasets; however, the differences are statistically significant (Mann–Whitney U test, $p < 0.0001$). The main differences between datasets are explained by VACV exhibiting a higher median (consistent with the observation in Figure~\ref{fig:plaque-results}(b)), and MAYV/CVB3 showing a more pronounced left tail, indicating a larger number of false positives.

Figure~\ref{fig:error} shows symmetric relative plaque-count error for wells containing plaques (GT > 0) and false-positive detections in empty wells. Both datasets exhibited right-skewed error distributions with most values < 0.2. MAYV/CVB3 showed a broader spread and slightly higher median than VACV (Fig.~\ref{fig:error}a–b), but the difference was not statistically significant (Mann–Whitney U, p = 0.38). The false-positive rate in empty wells was higher for MAYV/CVB3 (~0.74) than for VACV (0.50) (Fig.\ref{fig:error}c).

%To further characterise plaque-count discrepancies between datasets, Fig.~\ref{fig:error} shows the distribution of symmetric relative counting error, defined as |GT - pred| / (GT + pred), computed for wells containing plaques (GT > 0), together with the false-positive detection rate in empty wells.
%
%The symmetric relative error distributions for both datasets were right-skewed, with most wells concentrated at low error values (< 0.2) (Fig.~\ref{fig:error}, left). A partial overlap between datasets was observed, and MAYV/CVB3 showed a broader spread with more high-error values than VACV.
%
%The boxplot summary (Fig.~\ref{fig:error}, middle) showed a slightly higher median and wider interquartile range for MAYV/CVB3 than for VACV. However, a two-sided Mann–Whitney U test did not indicate a statistically significant difference between the relative error distributions (p = 0.38).
%
%Analysis of empty wells (Fig.~\ref{fig:error}, right) showed a higher false-positive detection rate for MAYV/CVB3 (~0.74) than for VACV (~0.50).

\subsection*{PFU/mL estimation results}

\begin{figure}[t]
    \centering
    \includegraphics[width=0.7\linewidth]{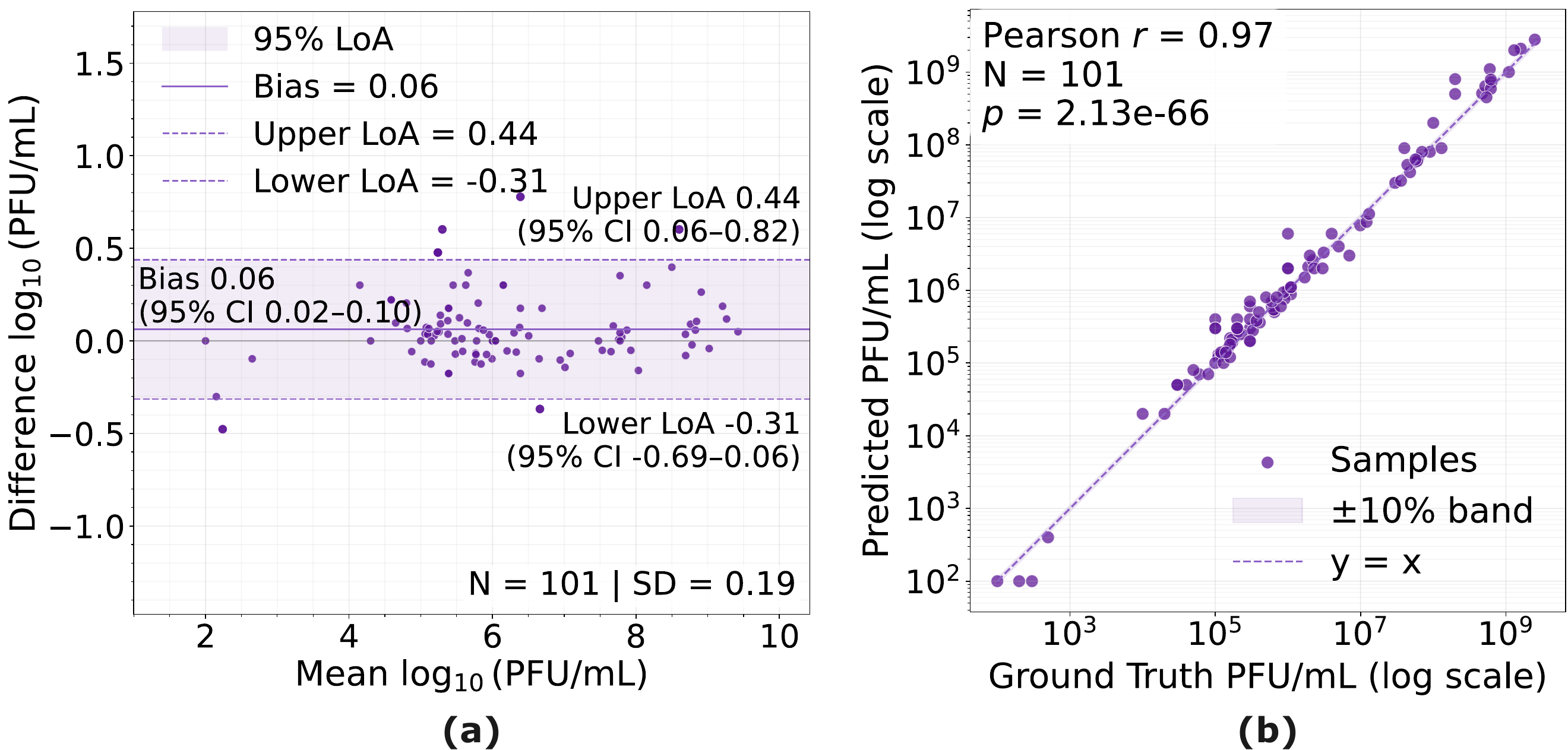}
    \caption{Comparison of PFU/mL values computed for the MAYV/CVB3 dataset using plaque counts obtained from manual annotation and automated plaque detection. a: Bland–Altman plot with agreement between the two PFU/mL measurements. b: Scatter plot comparing PFU/mL values derived from manual and automated plaque segmentations.}
    \label{fig:pfu}
\end{figure}

Figure~\ref{fig:pfu} presents a comparison between PFU/mL values computed from manual plaque annotations and those obtained using automated plaque segmentation for the MAYV/CVB3 dataset. The left panel shows a Bland–Altman plot assessing the agreement between PFU/mL values derived from the two plaque-counting methods. 174 out of 180 samples fall within the 95\% limits of agreement. The mean difference between the two measurements is $0.06$, with limits of agreement ranging from $-0.31$ to $0.44$. No evident trend is observed across the range of PFU/mL values. The right panel shows a scatter plot comparing PFU/mL values computed manually and automatically. A significant agreement is observed between both values (Pearson correlation coefficient $r = 0.975$, $N = 180$).

%Figure~\ref{fig:pfu} compares PFU/mL values derived from manual plaque annotations and automated segmentation for the MAYV/CVB3 dataset. The left panel presents a Bland–Altman analysis of agreement, with 174/180 samples within the 95\% limits of agreement. The mean difference was $0.06$, with limits ranging from $-0.31$ to $0.44$, and no systematic trend across PFU/mL values. The right panel shows a scatter plot demonstrating strong concordance between methods (Pearson $r=0.975$, $N=180$).

\section*{Discussion}

%We presented an end-to-end workflow for automated virus titration from plaque assay plate images based on machine learning. We validated the components of the proposed pipeline at multiple levels, including well segmentation across different plate types (with natural variation in acquisition angle and illumination), plaque-level segmentation and counting agreement with manual annotations, robustness across plaque-density regimes, and concordance of derived PFU/mL values with those computed from expert counts. Collectively, these results support the proposed workflow as a practical tool to assist plaque assay analysis in laboratory settings by improving reproducibility and reducing manual effort and inter-operator variability. Hence, we have released this pipeline as an open-source tool, \textit{Titra}, that we expect to enable virology researchers to improve their overall efficiency.

We presented an end-to-end machine learning workflow for automated virus titration from plaque assay images. The pipeline was validated at multiple levels, including well segmentation across diverse plate types and acquisition conditions, plaque segmentation and counting agreement with expert annotations, robustness across plaque-density regimes, and concordance of derived PFU/mL values with manual calculations. Together, these results demonstrate that the proposed workflow is a practical tool for laboratory plaque assay analysis, improving reproducibility while reducing manual effort and inter-operator variability. The pipeline will be released as the open-source tool \textit{Titra} upon acceptance of this manuscript to facilitate efficient and scalable viral quantification.
%The pipeline has been released as the open-source tool \textit{Titra} to facilitate efficient and scalable viral quantification.

%Accurate well detection is a prerequisite for reliable plaque quantification, as all downstream steps depend on correct identification and cropping of individual wells. In many existing open-source tools (e.g. ViralPlaque\cite{Cacciabue2019ViralPlaque}, Plaque2.0~\cite{Yakimovich2015Plaque20}), this step is often delegated to the user, which can be time-consuming and introduces user-dependent variability. By contrast, our workflow relies on a dedicated model that accurately identifies wells across multiple plate configurations (Figure~\ref{fig:well-segmentation-performance}). The main limitation observed is reduced performance for transparent wells as we observed in Figure~\ref{fig:well-segmentation-not-detected}. However, as transparent wells do not contain visible plaques and are not used for plaque counting or PFU estimation, this limitation can be handled within the pipeline without affecting the final biological readout. %Accordingly, transparent wells were excluded from the MAYV/CVB3 dataset prior to evaluation, given the high proportion of transparent wells in that dataset (31\%).

Accurate well detection is essential for reliable plaque quantification, as downstream steps depend on correct identification and cropping of individual wells. In many open-source tools (e.g. ViralPlaque\cite{Cacciabue2019ViralPlaque}, Plaque2.0~\cite{Yakimovich2015Plaque20}), this step is delegated to the user, which is time-consuming and introduces user-dependent variability. In contrast, our workflow uses a dedicated model that identifies wells across multiple plate configurations (Figure~\ref{fig:well-segmentation-performance}). Performance is reduced for transparent wells (Figure~\ref{fig:well-segmentation-not-detected}); however, as these wells contain no visible plaques and are not used for plaque counting or PFU estimation, this limitation does not affect the final biological readout.

%Plaque detection results indicate that the workflow provides reliable plaque quantification in most scenarios, despite variability in plaque size, contrast, and spatial distribution. Quantitative analyses (Figure~\ref{fig:plaque-results}) show strong agreement between automated and manual plaque counts, supported by high Pearson correlation coefficients and narrow limits of agreement in the Bland--Altman analysis. These findings suggest that automated counts remain consistent with manual annotations over a broad range of plaque densities.

Plaque detection results demonstrate reliable quantification across variations in plaque size, contrast, and spatial distribution. Quantitative analysis (Figure~\ref{fig:plaque-results}) shows strong agreement between automated and manual counts, reflected by high Pearson correlations and narrow Bland–Altman limits of agreement. These results indicate that automated counts remain consistent with expert annotations across a broad range of plaque densities.

A systematic pattern emerges at high plaque densities, particularly in the VACV dataset, where plaque overlap is more frequent (Figure~\ref{fig:plaque-results}(a)). In these cases, extensive confluence reduces the visibility of individual plaque boundaries, which can lead to under-segmentation and undercounting. This behaviour is reflected by increasingly negative differences in the Bland--Altman analysis at higher mean counts (Figure~\ref{fig:plaque-results}(b)) and by the qualitative examples, where closely overlapping plaques are difficult to disentangle (Figure~\ref{fig:plaque-results}(a)). Importantly, these high-density wells also increase ambiguity during manual counting, as indicated by the expert comparison (Figure~\ref{fig:comparison-with-experts}), and represent a limited subset of the evaluated data.

Expert comparison contextualises model errors relative to interobserver variability. In wells with low plaque counts, correlation between automated and expert annotations is high, and discrepancies are typically driven by ambiguous small plaques or plaque-like artefacts (Figure~\ref{fig:comparison-with-experts}). In such cases, interobserver variability can match the difference between automated and manual counts, indicating that part of the observed model “errors” lies within the range of human uncertainty. This aligns with qualitative examples showing inconsistent annotation of small plaques across experts.

%Despite plaque-level limitations under extreme confluence, PFU/mL estimation remains strongly concordant with manual calculations in MAYV/CVB3 (Figure~\ref{fig:pfu}), indicating that moderate counting discrepancies do not necessarily translate into biologically meaningful differences in virus titration. Unfortunately, a similar experiment cannot be produced in VACV because no data is provided regarding virus dilution factor. 

Despite plaque-level limitations under extreme confluence, PFU/mL estimates remain strongly concordant with manual MAYV/CVB3 calculations (Figure~\ref{fig:pfu}), indicating that moderate counting discrepancies do not necessarily affect virus titration. A similar experiment cannot be conducted for VACV because virus dilution factors are not provided.

%Relative plaque-count error was comparable between datasets for wells containing plaques, consistent with the non-significant distribution difference. In contrast, MAYV/CVB3 showed a markedly higher false-positive rate in empty wells, indicating more frequent detection of plaque-like structures in the absence of plaques. This suggests that plaque–background discrimination is more challenging in MAYV/CVB3 plates.

Relative plaque-count error was comparable between datasets for wells containing plaques, consistent with the non-significant distribution difference. However, MAYV/CVB3 showed a higher false-positive rate in empty wells, indicating more frequent detection of plaque-like structures and suggesting plaque–background discrimination is more challenging in MAYV/CVB3 plates.

%The distribution of relative errors (Figure~\ref{fig:error}) highlights dataset-dependent behaviour: MAYV/CVB3 shows a slight tendency towards overestimation, whereas VACV shows a tendency towards underestimation. This is a natural consequence of the overlapping phenomena detailed in previous paragraphs, and is expected to occur even with human labellers. %These findings support the use of relative error metrics when comparing performance across datasets with heterogeneous plaque-density regimes.

A key requirement for automated plaque analysis is generalisation across viruses, since plaque morphology can vary substantially in diameter, shape, spatial distribution, and density (Figure~\ref{fig:plaque-results}(a)). To assess this capability, we evaluated the workflow on assays from multiple viral species, including Mayaro virus, Coxsackievirus B3, and vaccinia virus (VACV). Across these datasets, we observed strong agreement with manual plaque counts despite pronounced differences in plaque appearance and concentration, supporting the use of the workflow as a virus-agnostic approach for automated plaque assay quantification. We encourage the research community to perform further experiments in other viral species using this tool. Furthermore, we invite computer science researchers to produce new models and integrate them in Titra, to keep extending its applicability.

Beyond standard plaque assay analysis, the modular design of the workflow makes it suited to adaptation for other virological assays that rely on plaque- or well-level readouts. For example, plaque reduction neutralisation tests (PRNT) require plaque enumeration across serum dilutions followed by downstream neutralisation calculations; the proposed workflow could automate the plaque counting step while preserving established downstream analyses. Similarly, $TCID_{50}$ assays rely on identifying wells with and without cytopathic effects rather than enumerating plaques; in this setting, the well detection and segmentation components can be reused, while the plaque-level module can be adapted for binary well classification. These extensions would require targeted validation but illustrate how the current pipeline can serve as a foundation for broader assay automation.

Despite the utility of the proposed approach, some limitations remain and motivate future work. First, as discussed before, wells with very high plaque density---particularly for viruses that frequently yield confluent patterns, such as VACV---remain challenging for both automated and manual analysis, and are associated with increased inter-expert variability (Figure~\ref{fig:comparison-with-experts}). In routine laboratory practice, such wells are often excluded because manual counting becomes unreliable \cite{payne2022viruses}. Improving performance in these regimes, for example, through explicit modelling of confluence or alternative counting strategies, is an important direction for further development. Second, expert annotators may exclude atypically sized plaque-like regions as artefacts, whereas the current implementation includes all detected plaque-like regions. Incorporating user-configurable filtering based on plaque size (or relative diameter distributions within a well) could better align automated outputs with expert counting practices in ambiguous cases, while preserving transparency and user control. Furthermore, this filtering closes a feedback loop that could be eventually exploited to retrain models and achieved better results.
The integration of the full pipeline within \textit{Titra} enables end-to-end analysis of plaque assay plates in a single automated run. On average, the combined processing time is approximately 30 s for well detection and an additional 30 s per well for plaque detection when executed without GPU acceleration. In local tests (i.e., without AWS infrastructure), we observed that well detection completes in approximately 4 s, with plaque segmentation requiring roughly 4 s per well. 
Importantly, the application also allows expert users to review and manually correct segmentation results when necessary, ensuring that automated outputs can be refined to reflect expert judgement in ambiguous cases.

\section*{Methods}

\subsection*{Proposed approach}

Plaque assay images can vary markedly across viruses, plate formats, and acquisition conditions, requiring segmentation methods that remain robust under heterogeneous appearances. We therefore adopted a foundation-model approach and used the SAM2 \cite{ravi2024sam} as the core segmentation engine. SAM2 extends the promptable segmentation paradigm introduced by SAM \cite{Kirillov2023SAM} and has demonstrated strong accuracy and efficiency across diverse image and video segmentation benchmarks \cite{ravi2024sam}. In parallel, SAM-based models have been shown to transfer effectively to biomedical imaging domains with challenging visual characteristics, including microscopy (\(\mu\)SAM) \cite{archit2025segment} and universal medical image segmentation (MedSAM) \cite{MedSAM}. Leveraging these properties, our pipeline treats well and plaque segmentation as prompt-guided instance segmentation tasks, enabling a modular design that can incorporate future model updates or domain-specific adaptation when required.

\subsubsection*{Well segmentation}

Well segmentation was performed in a zero-shot setting using SAM2 initialised from the pre-trained checkpoint \texttt{\url{sam2.1_hiera_base_plus.pt}} with configuration file \texttt{\url{sam2.1_hiera_b+.yaml}}. This variant was chosen to balance segmentation capacity and computational efficiency, and no fine-tuning was applied.

Prior to inference, input images were rescaled with preserved aspect ratio, setting the longest side to 256 pixels. Candidate masks were generated using the SAM2 automatic mask generator (\texttt{\url{points\_per\_side}}=12, \texttt{\url{min\_mask\_region\_area}}=100). Built-in post-processing was disabled in favor of a custom filtering stage.

Candidate masks were then filtered using geometric constraints to remove spurious detections and non-well regions. For each candidate region, we computed the area, perimeter, circularity ($C = 4\pi A / P^{2}$), and eccentricity. Only regions exceeding 100 pixels in area and exhibiting circularity greater than 0.7 were retained. For each validated mask, a bounding box was computed and mapped back to the original image resolution.

To determine the plate layout, the validated well masks were further processed to estimate the relative position of each well within the grid. This positional information was used to assign each well to its corresponding row and column, enabling correct association with the experimental dilution scheme during subsequent PFU/mL calculation.

This design prioritises robust and reliable well detection under CPU-only inference, ensuring scalability and compatibility with standard laboratory computing environments.

\subsubsection*{Plaque segmentation}

Plaque segmentation was performed using the image encoder of SAM \cite{Kirillov2023SAM}, which was used as a frozen feature extractor owing to its strong generalisation capabilities and its ability to capture fine-grained object boundaries. We used the \texttt{ViT-B} Vision Transformer backbone, keeping the encoder weights frozen throughout training to preserve the pre-trained visual representations learned from large-scale image data. This strategy---freezing a large pre-trained encoder and training only a task-specific decoder---has been explored successfully in prior work \cite{dong2024efficient,fan2025research}.

A lightweight convolutional decoder was attached to the frozen encoder to predict plaque segmentation masks. The decoder consists of a $3 \times 3$ convolutional layer followed by a ReLU activation and a final $1 \times 1$ convolution producing a single-channel output. Only the decoder parameters were optimised, enabling efficient training with fewer computational resources while retaining the representational capacity of the SAM encoder. The model outputs pixel-wise plaque probability maps.

Model training was implemented using the PyTorch Lightning framework~\cite{Falcon_PyTorch_Lightning_2019} to support reproducible and structured experimentation. Optimisation was performed using Adam optimiser \cite{Kingma2015AdamAM} and a binary cross entropy loss \cite{ruby2020binary} objective, with learning rates explored in the range $1 \times 10^{-5}$ to $1 \times 10^{-3}$. Training was conducted for up to 50 epochs with a batch size of 2 on a NVIDIA GeForce RTX 4090. Performance was monitored using intersection-over-union (IoU), Dice coefficient, and pixel accuracy on both training and validation sets.%, with experiment tracking performed using Weights \& Biases. When training on multiple datasets (MAYV/CVB3 and VACV), a custom \texttt{BalancedTwoDatasetSampler} was used to ensure equal dataset representation within each batch, mitigating potential dataset imbalance.
For multi-dataset training (e.g., MAYV/CVB3 and VACV), balanced mini-batch sampling was applied to mitigate dataset imbalance. Each batch contained 50\% randomly selected samples from each dataset. When dataset sizes differed, the number of batches per epoch was limited by the smaller dataset.

Input data consisted of $1024 \times 1024$ RGB images corresponding to individual wells extracted from multi-well assay plates, paired with binary GT masks. Images were resized to the target resolution and standardised using ImageNet statistics, while GT masks were processed to ensure binary consistency.
To improve generalisation across viruses and experimental conditions, we applied extensive data augmentation using a RandAugment-inspired pipeline \cite{cubuk2020randaugment, moris2024semi}. The number of augmentation operations per sample (\texttt{num\_ops} $\in \{2,4,6,8\}$) and the augmentation magnitude (\texttt{magnitude} $\in \{3,7,9,11\}$) were selected via hyperparameter exploration. Geometric transformations (horizontal and vertical flips, and random rotations) were applied synchronously to images and masks, whereas colour-based augmentations were applied to images only and included \texttt{AutoContrast}, \texttt{Equalize}, \texttt{Solarize}, \texttt{ColorJitter}, \texttt{Sharpness}, and \texttt{Posterize}, as implemented in Torchvision~\cite{torchvision2016}.

\begin{figure}
    \centering
    \includegraphics[width=.6\linewidth]{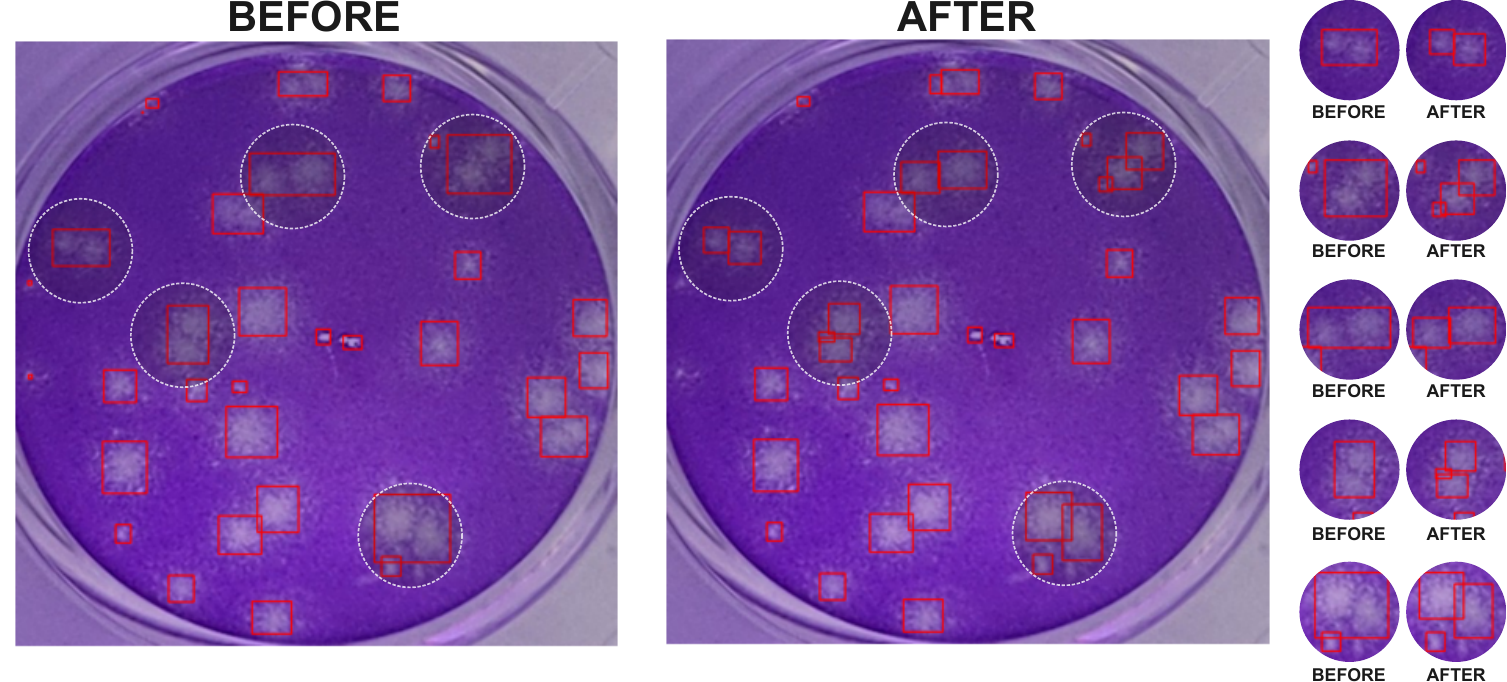}
    \caption{Post-processing refinement for separating overlapping plaques. Comparison of automated plaque detection before and after post-processing on a representative well. Red bounding boxes indicate detected plaques; white dashed circles highlight regions where overlapping or merged plaques were identified and subsequently separated. The right panel shows magnified examples illustrating the separation of overlapping plaques into distinct detections.}
    \label{fig:postprocessing}
\end{figure}

\subsubsection*{Plaque post-processing}

Since the model produces pixel-wise probability maps without explicit instance separation, post-processing was required to separate touching plaques and obtain accurate counts. The implemented strategy combines morphological separation, geometric validation, and density-based filtering  (Figure~\ref{fig:postprocessing}). 

The pipeline begins with watershed segmentation applied to the distance transform of the predicted binary mask, using local maxima as seed points to separate touching plaques. To validate plaque geometry and recover potentially merged regions, Hough circle detection was employed to identify approximately circular structures. Subsequently, a circularity analysis was used to flag irregular or ambiguous candidates based on a threshold of 0.5.
Subsequently, a DBSCAN clustering procedure~\cite{ester1996density} was applied to merge nearby peaks and reduce over-segmentation artefacts in dense plaque configurations. Finally, bounding boxes were delineated for each detected mask, and detections smaller than $10 \times 10$ pixels were assumed as noise and discarded. This combination of morphological, geometric, and density-based refinements enables robust plaque delineation under crowded and overlapping conditions.

\subsection*{Materials}

\subsubsection*{MAYV/CVB3 dataset}

The MAYV/CVB3 dataset was developed at the Instituto Pasteur de Montevideo and comprises 101 plaque assay images for training and validation, and 17 images for testing. All images were captured using a smartphone camera under ambient laboratory lighting, with the plates placed against a uniform white background to enhance contrast.

%For plaque assays, Vero cells (ATCC CCL-81) were seeded into 12-well and 6-well plates one day prior to infection at a density of $2 \times 10^{5}$ cells per well. Alphavirus stocks were serially diluted (ten-fold) in serum-free DMEM. Cells were washed twice with PBS and infected with $100~\mu\mathrm{L}$ each virus dilution for 1~h at $37^{\circ}\mathrm{C}$. Following infection, a semi-solid overlay consisting of DMEM supplemented with 2\% fetal bovine serum and 0.8\% agarose (Invitrogen) was added. After 45~h of incubation at $37^{\circ}\mathrm{C}$, cells were fixed with 4\% paraformaldehyde for 40~min. The agarose overlay was then removed, and each well was stained with 0.2\% crystal violet for 40~min. Plates were rinsed with tap water, air-dried, and imaged for plaque detection.

For plaque assays, Vero cells (ATCC CCL-81) were seeded in 12- and 6-well plates one day before infection at $2 \times 10^{5}$ cells per well. Mayaro virus and Coxsackievirus B3 stocks were serially diluted (ten-fold) in serum-free DMEM. Cells were washed twice with PBS and infected with $100~\mu\mathrm{L}$ of each dilution for 1h at $37^{\circ}\mathrm{C}$. A semi-solid overlay (DMEM supplemented with 2\% fetal bovine serum and 0.8\% agarose; Invitrogen) was then added. After 45h at $37^{\circ}\mathrm{C}$, cells were fixed with 4\% paraformaldehyde for 40min. The overlay was removed, wells were stained with 0.2\% crystal violet for 40min, rinsed with tap water, air-dried, and imaged for plaque detection.

Manual well and plaque segmentations were created using the open-source annotation tool Label Studio \cite{LabelStudio}. During dataset curation, a small subset of images was excluded from the training and validation sets due to excessive plaque density or fully transparent wells, which made accurate manual annotation infeasible. This filtering step aimed to ensure high-quality annotations and to prevent the inclusion of ambiguous cases that could introduce noise or bias during model training. Nonetheless, images containing partially visible or sparsely distributed plaques were retained to preserve dataset diversity and improve model generalisation. To avoid data leakage, images from the same experimental plate were kept within a single subset (training, validation, or test).

\subsubsection*{Vaccinia dataset}

The vaccinia virus (VACV) dataset is a publicly available resource introduced by De~\textit{et~al.}~\cite{De2025VACVPlaque}. It comprises 211 digital photographs of 6-well tissue culture plates used for plaque assays. Each image includes manually annotated masks for both wells and plaques. The dataset was divided into training, validation, and test subsets containing 149, 43, and 22 images, respectively.
The assays were performed on BSC40 cell monolayers infected with the VACV Western Reserve strain under various experimental conditions. Images were acquired using digital cameras.
Given the availability of high-quality manual annotations, we used this dataset both for direct model evaluation and to assess the generalisation ability of our approach across datasets with differing imaging conditions and biological setups.

\bibliography{ak-virus-titration}

\begin{thebibliography}{10}
\urlstyle{rm}
\expandafter\ifx\csname url\endcsname\relax
  \def\url#1{\texttt{#1}}\fi
\expandafter\ifx\csname urlprefix\endcsname\relax\def\urlprefix{URL }\fi
\expandafter\ifx\csname doiprefix\endcsname\relax\def\doiprefix{DOI: }\fi
\providecommand{\bibinfo}[2]{#2}
\providecommand{\eprint}[2][]{\url{#2}}

\bibitem{Masci2019VirPlaque}
\bibinfo{author}{Masci, A.~L.} \emph{et~al.}
\newblock \bibinfo{journal}{\bibinfo{title}{Integration of fluorescence detection and image-based automated counting increases speed, sensitivity, and robustness of plaque assays}}.
\newblock {\emph{\JournalTitle{Molecular Therapy -- Methods \& Clinical Development}}} \textbf{\bibinfo{volume}{14}}, \bibinfo{pages}{270--274}, \doiprefix\url{10.1016/j.omtm.2019.07.007} (\bibinfo{year}{2019}).

\bibitem{Yakimovich2015Plaque20}
\bibinfo{author}{Yakimovich, A.} \emph{et~al.}
\newblock \bibinfo{journal}{\bibinfo{title}{Plaque2.0--a high-throughput analysis framework to score virus-cell transmission and clonal cell expansion}}.
\newblock {\emph{\JournalTitle{PLOS ONE}}} \textbf{\bibinfo{volume}{10}}, \bibinfo{pages}{e0138760}, \doiprefix\url{10.1371/journal.pone.0138760} (\bibinfo{year}{2015}).

\bibitem{Katzelnick2018Viridot}
\bibinfo{author}{Katzelnick, L.~C.} \emph{et~al.}
\newblock \bibinfo{journal}{\bibinfo{title}{Viridot: An automated virus plaque (immunofocus) counter for the measurement of serological neutralizing responses with application to dengue virus}}.
\newblock {\emph{\JournalTitle{PLOS Neglected Tropical Diseases}}} \textbf{\bibinfo{volume}{12}}, \bibinfo{pages}{e0006862}, \doiprefix\url{10.1371/journal.pntd.0006862} (\bibinfo{year}{2018}).

\bibitem{Cacciabue2019ViralPlaque}
\bibinfo{author}{Cacciabue, M.} \emph{et~al.}
\newblock \bibinfo{journal}{\bibinfo{title}{Viralplaque: a {Fiji} macro for automated assessment of viral plaque statistics}}.
\newblock {\emph{\JournalTitle{PeerJ}}} \textbf{\bibinfo{volume}{7}}, \bibinfo{pages}{e7729}, \doiprefix\url{10.7717/peerj.7729} (\bibinfo{year}{2019}).

\bibitem{Trofimova2021PST}
\bibinfo{author}{Trofimova, E.} \& \bibinfo{author}{Jaschke, P.~R.}
\newblock \bibinfo{journal}{\bibinfo{title}{Plaque size tool: an automated plaque analysis tool for simplifying and standardising bacteriophage plaque morphology measurements}}.
\newblock {\emph{\JournalTitle{Virology}}} \textbf{\bibinfo{volume}{561}}, \bibinfo{pages}{1--5}, \doiprefix\url{10.1016/j.virol.2021.05.011} (\bibinfo{year}{2021}).

\bibitem{Phanomchoeng2022PeerJCS}
\bibinfo{author}{Phanomchoeng, G.} \emph{et~al.}
\newblock \bibinfo{journal}{\bibinfo{title}{Machine-learning-based automated quantification machine for virus plaque assay counting}}.
\newblock {\emph{\JournalTitle{PeerJ Computer Science}}} \textbf{\bibinfo{volume}{8}}, \bibinfo{pages}{e878}, \doiprefix\url{10.7717/peerj-cs.878} (\bibinfo{year}{2022}).

\bibitem{Liu2023NatBiomedEng}
\bibinfo{author}{Liu, T.} \emph{et~al.}
\newblock \bibinfo{journal}{\bibinfo{title}{Rapid and stain-free quantification of viral plaque via lens-free holography and deep learning}}.
\newblock {\emph{\JournalTitle{Nature Biomedical Engineering}}} \textbf{\bibinfo{volume}{7}}, \bibinfo{pages}{1451--1462}, \doiprefix\url{10.1038/s41551-023-01057-7} (\bibinfo{year}{2023}).

\bibitem{Kirillov2023SAM}
\bibinfo{author}{Kirillov, A.} \emph{et~al.}
\newblock \bibinfo{journal}{\bibinfo{title}{Segment anything}}.
\newblock {\emph{\JournalTitle{arXiv preprint arXiv:2304.02643}}} \doiprefix\url{10.48550/arXiv.2304.02643} (\bibinfo{year}{2023}).

\bibitem{De2025VACVPlaque}
\bibinfo{author}{De, T.} \emph{et~al.}
\newblock \bibinfo{journal}{\bibinfo{title}{A digital photography dataset for {Vaccinia} virus plaque quantification using deep learning}}.
\newblock {\emph{\JournalTitle{Scientific Data}}} \textbf{\bibinfo{volume}{12}}, \bibinfo{pages}{719}, \doiprefix\url{10.1038/s41597-025-05030-8} (\bibinfo{year}{2025}).

\bibitem{Emi2025PlaQuest}
\bibinfo{author}{Emi, A.} \emph{et~al.}
\newblock \bibinfo{journal}{\bibinfo{title}{Development of an automated plaque-counting program for the quantification of the {Chikungunya} virus}}.
\newblock {\emph{\JournalTitle{Scientific Reports}}} \textbf{\bibinfo{volume}{15}}, \bibinfo{pages}{12429}, \doiprefix\url{10.1038/s41598-025-97590-3} (\bibinfo{year}{2025}).

\bibitem{ravi2024sam}
\bibinfo{author}{Ravi, N.} \emph{et~al.}
\newblock \bibinfo{journal}{\bibinfo{title}{Sam 2: Segment anything in images and videos}}.
\newblock {\emph{\JournalTitle{arXiv preprint arXiv:2408.00714}}}  (\bibinfo{year}{2024}).

\bibitem{kirillov2023segment}
\bibinfo{author}{Kirillov, A.} \emph{et~al.}
\newblock \bibinfo{title}{Segment anything}.
\newblock In \emph{\bibinfo{booktitle}{Proceedings of the IEEE/CVF international conference on computer vision}}, \bibinfo{pages}{4015--4026} (\bibinfo{year}{2023}).

\bibitem{nguyen2022trustworthy}
\bibinfo{author}{Nguyen, T. T.~D.} \emph{et~al.}
\newblock \bibinfo{journal}{\bibinfo{title}{How trustworthy are performance evaluations for basic vision tasks?}}
\newblock {\emph{\JournalTitle{IEEE Transactions on Pattern Analysis and Machine Intelligence}}} \textbf{\bibinfo{volume}{45}}, \bibinfo{pages}{8538--8552} (\bibinfo{year}{2022}).

\bibitem{payne2022viruses}
\bibinfo{author}{Payne, S.}
\newblock \emph{\bibinfo{title}{Viruses: from understanding to investigation}} (\bibinfo{publisher}{Elsevier}, \bibinfo{year}{2022}).

\bibitem{archit2025segment}
\bibinfo{author}{Archit, A.} \emph{et~al.}
\newblock \bibinfo{journal}{\bibinfo{title}{Segment anything for microscopy}}.
\newblock {\emph{\JournalTitle{Nature Methods}}} \textbf{\bibinfo{volume}{22}}, \bibinfo{pages}{579--591} (\bibinfo{year}{2025}).

\bibitem{MedSAM}
\bibinfo{author}{Ma, J.} \emph{et~al.}
\newblock \bibinfo{journal}{\bibinfo{title}{Segment anything in medical images}}.
\newblock {\emph{\JournalTitle{Nature Communications}}} \textbf{\bibinfo{volume}{15}}, \bibinfo{pages}{654} (\bibinfo{year}{2024}).

\bibitem{dong2024efficient}
\bibinfo{author}{Dong, G.} \emph{et~al.}
\newblock \bibinfo{journal}{\bibinfo{title}{An efficient segment anything model for the segmentation of medical images}}.
\newblock {\emph{\JournalTitle{Scientific Reports}}} \textbf{\bibinfo{volume}{14}}, \bibinfo{pages}{19425} (\bibinfo{year}{2024}).

\bibitem{fan2025research}
\bibinfo{author}{Fan, K.} \emph{et~al.}
\newblock \bibinfo{journal}{\bibinfo{title}{Research on medical image segmentation based on sam and its future prospects}}.
\newblock {\emph{\JournalTitle{Bioengineering}}} \textbf{\bibinfo{volume}{12}}, \bibinfo{pages}{608} (\bibinfo{year}{2025}).

\bibitem{Falcon_PyTorch_Lightning_2019}
\bibinfo{author}{Falcon, W.} \& \bibinfo{author}{{The PyTorch Lightning team}}.
\newblock \bibinfo{title}{{PyTorch Lightning}}, \doiprefix\url{10.5281/zenodo.3828935} (\bibinfo{year}{2019}).

\bibitem{Kingma2015AdamAM}
\bibinfo{author}{Kingma, D.~P.} \& \bibinfo{author}{Ba, J.}
\newblock \bibinfo{title}{Adam: A method for stochastic optimization}.
\newblock In \emph{\bibinfo{booktitle}{International Conference on Learning Representations (ICLR)}} (\bibinfo{year}{2015}).

\bibitem{ruby2020binary}
\bibinfo{author}{Ruby, U.}, \bibinfo{author}{Yendapalli, V.} \emph{et~al.}
\newblock \bibinfo{journal}{\bibinfo{title}{Binary cross entropy with deep learning technique for image classification}}.
\newblock {\emph{\JournalTitle{Int. J. Adv. Trends Comput. Sci. Eng}}} \textbf{\bibinfo{volume}{9}} (\bibinfo{year}{2020}).

\bibitem{cubuk2020randaugment}
\bibinfo{author}{Cubuk, E.~D.} \emph{et~al.}
\newblock \bibinfo{title}{Randaugment: Practical automated data augmentation with a reduced search space}.
\newblock In \emph{\bibinfo{booktitle}{Proceedings of the IEEE/CVF conference on computer vision and pattern recognition workshops}}, \bibinfo{pages}{702--703} (\bibinfo{year}{2020}).

\bibitem{moris2024semi}
\bibinfo{author}{Moris, E.} \emph{et~al.}
\newblock \bibinfo{title}{Semi-supervised learning with noisy students improves domain generalization in optic disc and cup segmentation in uncropped fundus images}.
\newblock In \emph{\bibinfo{booktitle}{Medical Imaging with Deep Learning}} (\bibinfo{year}{2024}).

\bibitem{torchvision2016}
\bibinfo{author}{TorchVision-maintainers}.
\newblock \bibinfo{title}{Torchvision: Pytorch's computer vision library} (\bibinfo{year}{2016}).

\bibitem{ester1996density}
\bibinfo{author}{Ester, M.} \emph{et~al.}
\newblock \bibinfo{title}{A density-based algorithm for discovering clusters in large spatial databases with noise.}
\newblock In \emph{\bibinfo{booktitle}{kdd}}, vol.~\bibinfo{volume}{96}, \bibinfo{pages}{226--231} (\bibinfo{year}{1996}).

\bibitem{LabelStudio}
\bibinfo{author}{Tkachenko, M.}, \bibinfo{author}{Malyuk, M.}, \bibinfo{author}{Holmanyuk, A.} \& \bibinfo{author}{Liubimov, N.}
\newblock \bibinfo{title}{{Label Studio}: Data labeling software} (\bibinfo{year}{2020-2025}).
\newblock \bibinfo{note}{Open source software available from https://github.com/HumanSignal/label-studio}.

\end{thebibliography}

\section*{Acknowledgements}
 We thank Mercedes Paz, Natalia Echeverría, Álvaro Fajardo, Paula Perbolianachis, and Juan Gandioli for their participation in testing Titra and for their valuable feedback during its evaluation.
 
\section*{Funding}
This work was funded by ANII (Uruguay), grant ART X 2024 1 180046, and by Arionkoder Global LLC.

\section*{Author contributions statement}

All authors contributed to the conceptualization and design of the solution. 
E.M. and J.I.O. designed, implemented and validated deep learning models. 
L.L.L., M.V., A.E.V., J.R., I.M. designed and implemented the web application.
A.C., S.R., I.F. and J.H. conducted the viral essay experiments and provided both image data and their annotations.
E.M., A.C., P.M., G.M. and J.I.O. analysed the results
E.M. and J.I.O. wrote the manuscript.
All authors reviewed the manuscript. 

\section*{Additional information}

\textbf{Data availability}\\
\noindent The VACVPlaque benchmark images used in this study are openly available on RODARE~\cite{De2025VACVPlaque} under a CC-BY 4.0 license. The remaining primary images were collected at Institut Pasteur de Montevideo under institutional agreements; these raw images will be published by the time of publication. %to qualified researchers upon reasonable request to the corresponding author, subject to approval by Institut Pasteur de Montevideo and execution of an appropriate data/material transfer agreement.
Derived annotations (well and plaque masks) and aggregate titration results required to reproduce our analyses will be publicly released upon acceptance of this manuscript. 

\subsection*{Code availability}
The Titra platform and the end-to-end analysis pipeline will be released as open-source software at \url{github.com/arionkoder/titra-ai} upon acceptance of this paper.

\noindent \textbf{Competing interests}
\noindent The authors declare no competing interests.

\end{document}